\documentclass[conference]{IEEEtran}
\ifCLASSINFOpdf
  
\else
\fi
\usepackage[pdftex]{graphicx}
\usepackage{amsmath}
\usepackage{microtype}
\usepackage{textcomp}
\usepackage{multicol}
\usepackage{float}
\usepackage{tikz}
\usetikzlibrary{arrows,positioning,shapes.geometric}
\IEEEoverridecommandlockouts 
\newcount\colveccount
\newcommand*\colvec[1]{
        \global\colveccount#1
        \begin{pmatrix}
        \colvecnext
}
\def\colvecnext#1{
        #1
        \global\advance\colveccount-1
        \ifnum\colveccount>0
                \\
                \expandafter\colvecnext
        \else
                \end{pmatrix}
        \fi
}

\begin{document}
%
\title{Traffic Sign Classification Using Deep Inception Based Convolutional Networks}

\author{\IEEEauthorblockN{Mrinal Haloi}
\IEEEauthorblockA{IIT Guwahati$ ^1$\\mrinal.haloi11@gmail.com}
\thanks{$ ^1$Indian Institute of Technology, Guwahati.\
\textcopyright~2015 Mrinal Haloi}
}


%


\maketitle

\begin{abstract}
In this work, we propose a novel deep network for traffic sign classification that achieves outstanding performance on GTSRB surpassing all previous methods. Our deep network consists of spatial transformer layers and a modified version of inception module specifically designed for capturing local and global features together. This features adoption allows our network to classify precisely intraclass samples even under deformations. Use of spatial transformer layer makes this network more robust to deformations such as translation, rotation, scaling of input images. Unlike existing approaches that are developed with hand-crafted features, multiple deep networks with huge parameters and data augmentations, our method addresses the concern of exploding parameters and augmentations. We have achieved the state-of-the-art performance of 99.81\% on GTSRB dataset.

Keywords: Advanced Driver Assistant System, Image Classification, Traffic Sign, Deep Networks

\end{abstract}


%
\IEEEpeerreviewmaketitle

\section{Introduction}
\begin{figure}
  \centering
      \includegraphics[width=3.3in,height=2.0in]{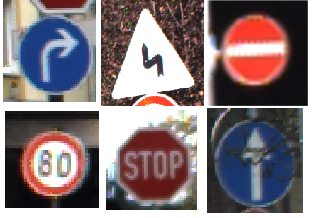}
\caption{Sample images of GTSRB dataset}
\label{fig:intro}
\end{figure}

Traffic signs classification is one of the foremost important integral parts of autonomous vehicles and advanced driver assistance systems (ADAS) \cite{haloi1, trivedisurvey, haloi2, haloi3, activity}. Most of the time driver missed traffic 
signs due to different obstacles and lack of attentiveness. Automating the process of classification of the traffic signs would help reducing accidents.
Traditional computer vision and machine learning based methods were widely used for traffic signs 
classification \cite{le,plsa}, but those methods were soon replaced by deep learning based classifiers. Recently deep convolutional networks have surpassed traditional learning methods in traffic signs classification. With the rapid advances of deep learning algorithm structures and feasibility of its high performance implementation with graphical processing units (GPU), it is advantageous to relook the traffic signs classification problems from the efficient deep learning perspective.       
 Classification of traffic signs is not so simple task, images are effected to adverse variation due to illumination, orientation, the speed variation of vehicles etc. Normally wide angle camera is mounted on the top of a vehicle to capture traffic signs and other related visual features for ADAS. This images are distorted due to several external factors including vehicles speed, sunlight, rain etc. Sample images from GTSRB dataset are shown in Fig.~\ref{fig:intro}.

 In this work, we have developed a new system for classification on the top of existing deep learning methods. We aim to address the traditional hand crafted data augmentation problem and also the reduction of an overwhelming number of parameters need to be learned by the system. With the reduction of a number of parameters, lesser number of computations and memory will be used. Instead of using a single sized filter for one convolutional layer, we would use multiple sized filters and concatenate those convolutional filters response to get more abstract representations in one layer. Also, there are disadvantages of using hand crafted data augmentation on some specific datasets for learning parameters that would be used for general purpose. Hence learning parameters for data modification inside network is a far better way to improve the accuracy of classification.   
 A modified inception module suitable for traffic sign classification is proposed to use with GoogLeNet \cite{googlenet} and to get way with traditional data augmentation we have incorporated spatial transformer network \cite{stn} layer with our proposed network. In terms of accuracy and number of parameters, this method has surpassed state of the art method for traffic sign classification.

Traffic sign classification becomes a mature area with the increasing focus on autonomous driving research. Notable research work exists on detection and classification traffic signs for advanced driver assistance systems. Most of the works attempted to address the challenged involved in
real life problems due to scaling, rotation, blurring etc. We will go through the overview  of  some relevant works since it is not possible to discuss all those research works. Most  of  the  works based on computer vision and machine learning algorithms which use data from several camera sensors mounted on the car roof at different angles. In some of the work, researchers explore
detection based on colour features, such as converting the colour space from RGB to HSV and then using colour thresholding method for detection and classification by using support vector  machine. In colour thresholding approach morphological operation like connected component
analysis was done for accurate location. Bahlmann et al\cite{bah} have used colour, shape, motion information and haar wavelet based features for detection, classification of the traffic sign. By using SVM based colour classification on a block of pixels Le et al \cite{le} addressed the problems of weather variation. German Traffic Sign Recognition Benchmark (GTSRB) is one of the reliable datasets for testing and validating traffic sign classification and detection algorithms. In the competition of GTSRB, top-performing algorithm exceeds best human classification accuracy. By using committee of neural networks Ciresan et al \cite{committe} achieved highest ever
performance of 99.46\%, which surpassed the best human performance of 98.84\%. Their proposed committee composed of 25 networks each having 3 convolutional and 2 fully connected networks with traditional data augmentations and jittering. The main disadvantages of this committe are multiples networks, a huge number of parameters ( around 90Millions) and dataset dependent handcrafted augmentations. Sermanet et al. proposed multi-scale convolutional network \cite{multi} with 2 different features stages, which has achieved 98.31\% accuracy in this dataset. In our previous work \cite{plsa} a probabilistic latent semantic analysis based model was proposed, which was built upon traditional handcrafted features extraction methods. Also, other algorithm based on k-d trees and random forest \cite{rf} achieved significant accuracy.

\section{Method}
Traffic signs classification are affected due to contrast variation, rotational and translational changes. It is possible
to nullify the effect of spatial transformations in an image undergo due to varying speed of vehicles camera by using multiple
transformations to the input image. But these handcrafted transformations are not effective always and vary with scenarios.
In this work, a spatial transformer network \cite{stn} capable of generating automatic transformation of input image is used to
make classification more robust and accurate along with a modified version of GoogLeNet \cite{googlenet}.

\subsection{Transformation invariant}
\begin{figure}
  \centering
      \includegraphics[width=3.3in,height=2.0in]{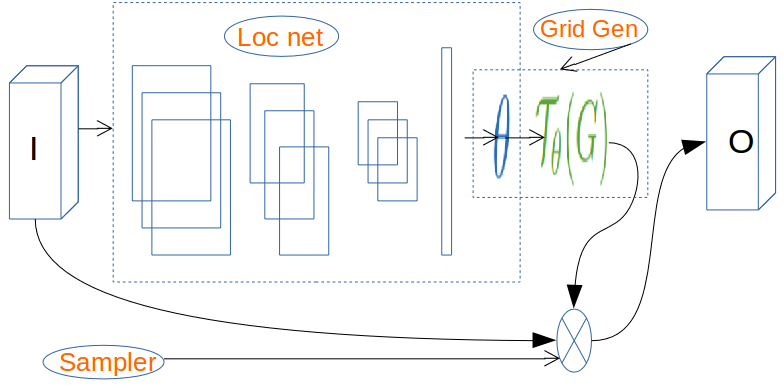}
\caption{Transformer Layer \cite{stn}}
\label{fig:loc}
\end{figure}

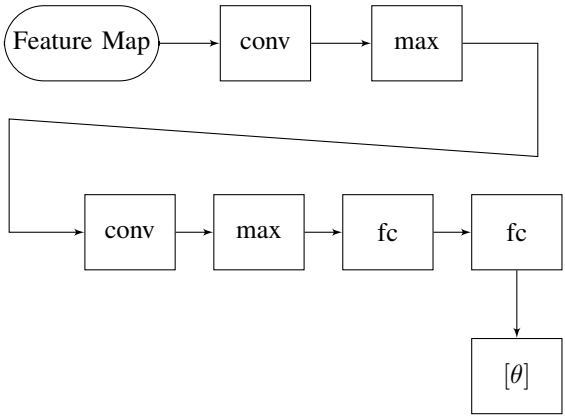
\begin{figure}
\begin{tikzpicture}[>=latex']
        \tikzset{block/.style= {draw, rectangle, align=center,minimum width=1.2cm,minimum height=1cm},
        rblock/.style={draw, shape=rectangle,rounded corners=1.5em,align=center,minimum width=2cm,minimum height=1cm},
        input/.style={ 
        draw,
        trapezium,
        trapezium left angle=60,
        trapezium right angle=120,
        minimum width=1cm,
        align=center,
        minimum height=.5cm
    },
        }
        \node [rblock]  (start) {Feature Map};
        \node [block, right =.8cm of start] (acquire) {conv};
        \node [block, right =.8cm of acquire] (rgb2gray) {max};
        \node [block, below right =1.5cm and -1cm of start] (otsu) {conv};
        \node [block, right =.5cm of otsu] (gc) {max};
        \node [block, right =.5cm of gc] (gchannel) {fc};
        \node [block, right =.5cm of gchannel] (closing) {fc};
        \node [block, below =.9cm of closing] (out) {$ [\theta] $};
        \node [coordinate, below right =1cm and 1cm of rgb2gray] (right) {};
        \node [coordinate,above left =1cm and 1cm of otsu] (left) {};

        \path[draw,->] (start) edge (acquire)
                    (acquire) edge (rgb2gray)
                    (rgb2gray.east) -| (right) -- (left) |- (otsu)
                    (otsu) edge (gc)
                    (gc) edge (gchannel)
                    (gchannel) edge (closing)
                    (closing) edge (out)
                    ;
 \end{tikzpicture}
       \caption{Localisation Network}
       \label{fig:locnet}
\end{figure}

Due to a moving camera, a image undergoes deformation like blurring, translational deformation, rotational
and scale deformations, skew etc. For classification feature map (include input image batch) would be passed through layers of 
spatial transformers. Spatial transformations modules are differentiable hence could be used with 
backpropagation algorithm for training. Spatial transformer layers consist of three parts such as 
localisation network, grid generator and the sampling unit. Figure~\ref{fig:loc} shows spatial transformer network with its components. This layer could be inserted at any point 
into the CNN network and it is efficient to deal with due to its very low computational overhead. Using
this layer with CNN obviated the use of handcrafted data augmentation such as translation, rotation etc.
and allows the network to learn active transformation of features map. Localisation network can be fully connected or convolutional 
neural network with one mandatory final regression layer to generate parameters $\theta$. Dimension of 
$ \theta $ depends on the parameterized transformation $ \tau_{\theta} $ type. To compute parameters 
localisation network may take input image or input feature map $ U \in \Re^{H \times W \times C} $ with 
$ W, H, C $ are the width, height and channels respectively. Localisation network can deal with multiples 
channels. Also localisation network Fig.~\ref{fig:locnet} may take any number of convolutional and fully connected layers as per application requirement. Using parameters produced by localisation network, grid generator creates set of points known 
as sampling grid. Sampling grid and input feature map (or input image) is used by a sampler to generate 
the transformed output map. Each pixel of the output feature map is computed using sampling kernel
centred at a definite input feature map location.
\begin{equation}
\label{eq:relation}
	\begin{aligned}
		\theta = f_{loc}(U) \\
		U \in \Re^{H \times W \times C}
	\end{aligned}
\end{equation}
For a input feature map pixel $ (x_{i}, y_{i})$ and learned 2D affine transformation parameters $ \tau_{\theta} $, the output 
feature map pixel $ (x_{i}^{o}, y_{i}^{o})$ is computed as follows
\begin{equation}
\label{eq:affine}
	\begin{aligned}
		\colvec{2}{x_{i}}{ y_{i}} =  \tau_{\theta}\colvec{3}{x_{i}^{o}}{y_{i}^{o}}{1}\\
		\tau_{\theta} = 
		\begin{bmatrix}
			\theta_{11} & \theta_{12} & \theta_{13}\\
			\theta_{21} & \theta_{22} & \theta_{23}
		\end{bmatrix}
	\end{aligned}
\end{equation}
Using the transformation defined in \eqref{eq:affine} we can perform operations such as translation, rotation, scale, skew and cropping in the input feature map. Parameters $ \theta$ are computed using equation \eqref{eq:relation}. Interestingly this transformation requires only 6 parameters to be learned by localisation network.

\begin{figure}
  \centering
      \includegraphics[width=3.3in,height=2.0in]{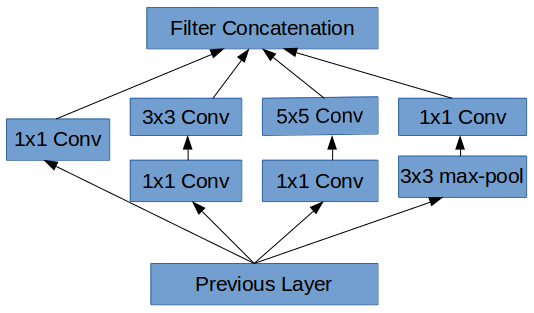}
\caption{Google Proposed Inception module}
\label{fig:gincept}
\end{figure}

\begin{figure}
  \centering
      \includegraphics[width=3.3in,height=2.0in]{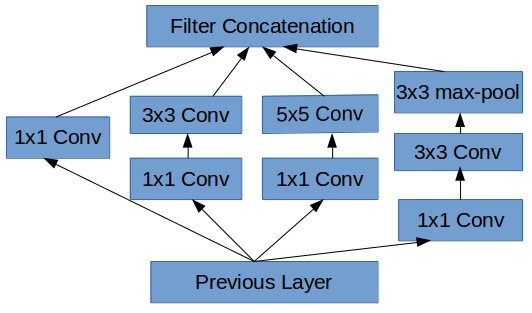}
\caption{Our Modified Inception module for Traffic Signs Classification}
\label{fig:myincept}
\end{figure}

\subsection{Proposed Pipeline}
A modified version of GoogLeNet \cite{googlenet} with batch normalization \cite{batch} is used as parent network for the classification task. GoogLeNet is based on the Inception architecture. Several Inception modules stacked upon each other to produce the final output. At the inception module varied size of convolutional filters were use to capture features of different abstraction. High level of abstraction is captured with filters of higher size and that of a lower level using small size filters. Processing
visual information at different scales and aggregating them result in an efficient level of abstraction. Since directly applying more
convolutional filters with image data and concatenating them is computationally expensive, so in the final Inception model a dimensionality reduction filters was used very applying abstraction level filters. For dimensionality reduction $ 1 \times 1 $ convolutional filters are used. Besides being very successful for dimensionality reduction, this filters also come to be useful as rectified linear activation. Inception architecture is efficient in terms of computational complexity with respect to number of units at each stage. 
 For our classification task a modified version of Inception module is used. For traffic sign classification local abstract features play important role. Signs belonging to same group have slight difference in local structure with each other, which make it hard to distinguish. A extra $ 3 \times 3 $ convolutional reduction kernel is added with max pooling at the top of it to capture discriminative local structure at the beginning itself. Signs belonging to different groups has global abstraction which can be captured using $ 5 \times 5 $ convolutional reduction kernel. Improved performance is observed with this architecture over normal Inception module. Figure~\ref{fig:gincept} shows Google inception module and Fig.~\ref{fig:myincept} shows our proposed inception module.

\begin{figure*}
  \centering
      \includegraphics[width=6.3in,height=2.0in]{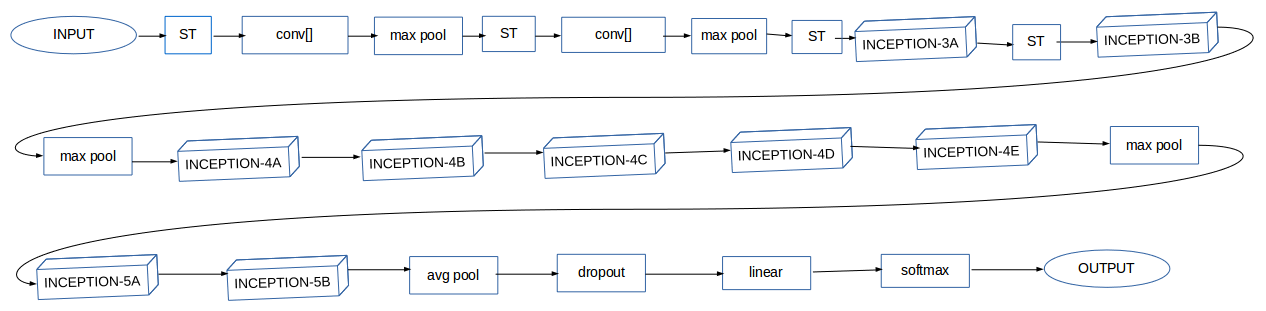}
\caption{Proposed network with Modified Inception}
\label{fig:network}
\end{figure*}

\begin{table*}[t]
\caption{Detailed description of deep networks parameters}
\label{tab:network}
  \centering
  \begin{tabular}{|c|c|c|c|c|c|c|c|c|c|c|}
\hline
 type & RF size/stride & output size & \#1x1 & \#3x3 reduce & \#3x3 & \#5x5 reduce & \#5x5 & \#3x3 reduce pool & \#3x3 & params \\
 \hline
 \hline
 ST1 & & & & & & & & & & 3M \\
 \hline
 conv & 5x5/2 &64x64x64 &  &   &   &   &   &  & & 1.6K \\
 \hline
 max-pool & 3x3/2 & 32x32x64&  &   &   &   &   & &  &  \\
 \hline
 ST2 & & & & & & & & & & 891K\\
 \hline
 conv & 3x3/1 &  32x32x192 &  &    &  192 &   &   &  & & 110K \\
 \hline
 max-pool & 3x3/2 &  16x16x192  &  &   &   &   &   &  & &  \\
 \hline
 ST3 & & & & & & & & & & 1M\\
 \hline
 mIncept(3a) &  &  16x16x288  & 64  & 96  & 128  & 16  & 32  & 64 & 64 & 206K \\
 \hline
 ST3 & & & & & & & & & & 1M\\
 \hline
 mIncept(3b) & &   16x16x480 &  128 & 128  & 192  & 32  & 96  & 64 & 64 & 436K \\
 \hline
 max-pool & 3x3/2 &  8x8x480 &  &   &   &   &   &   & & \\
 \hline
 mIncept(4a) &  &  8x8x512  & 192  & 96  & 208  & 16  & 48  & 48 & 64 & 395K \\
 \hline
 mIncept(4b) &  &    8x8x512 & 160 & 112  & 224  & 24  & 64  & 48 & 64 & 467K \\
 \hline
 mIncept(4c) &  &    8x8x512 & 128 & 128  & 256  & 24  & 64  &  64& 64& 546K \\
 \hline
 mIncept(4d) &  &    8x8x528 & 112 & 144  & 288  & 32  & 64  & 48 & 64 & 624K \\
 \hline
 mIncept(4e) &  &   8x8x832 & 256 & 160  &  320 & 32  & 128  & 48 & 128 & 880K \\
 \hline
 max-pool & 3x3/2 & 4x4x832   &  &   &   &   &   &  & &  \\
 \hline
 mIncept(5a) & &  4x4x832 & 256  & 160  & 320  & 32  & 128 & 48& 128& 965K \\
 \hline
 mIncept(5b) &  &    4x4x1024 & 320 & 192  & 320  & 48  & 128  & 32 & 256 & 1.2M \\
 \hline
 avg-pool & 4x4/1 &  1x1x1024  &  &   &   &   &   &  & &  \\
 \hline
 dropout &  &  1x1x1024   &  &   &   &   &   &  & &  \\
 \hline
 linear &  &  1x1x43  &  &   &   &   &   &  &  & 44K \\
 \hline
 softmax &  &  1x1x43  &  &   &   &   &   &  &  &  \\

\hline
 \end{tabular}
\end{table*}

\section{Results and Discussions}
We extensively evaluate our proposed deep networks on GTSRB (German Traffic Sign Recognition Benchmark) \cite{gtsrb} using our modified networks and also with original GoogLeNet. GTSRB is the standard state-of-the-art revelation benchmark for traffic sign recognition/classification. There are significant similarities of German traffic sign with other European countries and with Indian conventions, which make it suitable to explore. 
\subsection{Training and Datasets}
The proposed network was trained and tested using the machine learning library Torch \cite{torch} and two NVIDIA Tesla K40c GPU. For the implementation of the spatial transformer network, stn \cite{stnp} package was used.

For training and testing, GTSRB dataset contains 51839 images in 43 classes. We have selected 39,209 images for training and rest for testing. Images with deformation due to viewpoint variation, occlusion due to obstacles like trees, building etc., natural degrading, weather condition are considered in this dataset. We have resized all input images to $ 128 \times 128 $ using cubic interpolation method. 

For training, we have used SGD with momentum, with minibatch size of 20 images and learning rate of 0.00032. Dropout (40\%) was used for the fully connected layer. For SGD we have used momentum 0.9 with weight decay of 0.0918. Also, it has been observed that learning rate primarily influence training process. For activation Parametric Rectified Linear Unit (PReLU) \cite{prelu} is used. Instead of using parameter free ReLU, we have used PReLU for better accuracy. Parameters of PReLU are learned during training of the network. Also, networks weights are initialized using MSRA \cite{prelu} methods, which proved to be useful for PReLU activation unit based networks.
In Table~\ref{tab:network} a detailed description of proposed network is given. Number of $ 1 \times 1$ filters used for dimensionality reduction before $ 3 \times 3$ and $ 5 \times 5$ convolutions are referred as ``\#$ 3\times 3$ reduce" and ``\#$ 5\times 5$ reduce". Also ``\#$ 3\times 3$ reduce pool" refers to number of $ 1 \times 1$ filters used before $ 3 \times 3$ convolution and max pooling. In addition to that in Table~\ref{tab:network} $mIncept(..) $ refers to our proposed inception module. It's notable that apart from the inception module, proposed method primary network also have slight difference from original GoogLeNet. The proposed network is 21 layers deep including only the layers with parameters, excluding pooling layers and spatial transformer layers (have parameters). Since spatial transformer layer (ST) is a features transformer layer with its own network parameters, it doesn't impact the features learning process of rest of the network. If pooling layers are included, then our $ mIncept(..)$ module would have a depth of 3. Including pooling and ST the network is 39 layers deep. We have used four spatial transformer layers (networks), two of them before two convolutional layers and other two before modified inception modules. Also the network configurations of spatial transformer layers ST1, ST2 and ST3 are different and detailed information is given in Table~\ref{tab:stn}.

\begin{table}[h]
\caption{Spatial Transformer layer(Networks) configurations}
\label{tab:stn}
  \begin{center}
  \begin{tabular}{|c|c|c|c|c|c|c|}
\hline
Net & conv/stride & max-pool & conv/stride & max-pool & fc & fc \\
\hline
ST1 & 128,5x5/2 & yes & 192, 5x5/2 & yes & 192 & 192 \\
\hline
ST2 & 128, 5x5/2 & yes & 192, 5x5/2 & no & 192 & 192 \\
\hline
ST3 & 128, 3x3/2 & no & 192, 3x3/1 & yes & 192 & 192 \\
\hline
 \end{tabular}
 \end{center}
\end{table}

\begin{table*}[t]
\caption{Comparions of groups accuracy (Top-1) with state-of-the-arts}
\label{tab:group}
  \centering
  \begin{tabular}{|c|c|c|c|c|c|c|}
\hline
 Algorithm & Speed Limits & Prohibitions & Derestrictions & Mandatory & Danger & Unique\\
 \hline
 \hline
\textbf{Proposed Method} & 99.86 & 100 & 99.95 & 99.72 & 99.89 & 99.87\\
\hline
Committee of CNNs \cite{committe} & 99.47 & 99.93 & 99.72 &99.89& 99.07 & 99.22\\
\hline
Human \cite{gtsrb} & 98.32 & 99.87 & 98.89 & 100.00 & 99.21 & 100.00\\
\hline
Multi-Scale CNN \cite{multi} & 98.61 & 99.87 & 94.44 & 97.18 & 98.3 & 98.63\\
\hline
pLSA \cite{plsa} \ & 98.82 & 98.27 & 97.93 & 96.86 & 96.95 & 100.00\\
\hline
Random Forest \cite{rf} & 95.95 & 99.13 & 87.50 & 99.27 & 92.08 & 98.73\\
\hline
 \end{tabular}
\end{table*}

\subsection{Comparions with State-of-the-Art}
This method has several advantages over existing state of the art methods in terms of performance, scalability and memory requirement. Recent high performed method Committee of CNNs\cite{committe} have used 25 networks with 3 convolutional layers and 2 fully connected layers along with manual data augmentation. On original dataset they have modified each image using translation, rotation etc., to get five modified version of that image. Committee of CNNs end up with total around ~90 Million parameters whereas, in our method, we have around 10.5 Million parameters. \\
 Overall accuracy comparisons with different high performing approaches are shown in Table~\ref{tab:accuracy}. We have also reported the accuracy of our deep networks with the Google inception module, which is slightly lower than the accuracy obtained using our modified inception module.\\
GTSRB dataset composed of mainly 6 high-level groups. Classification of images belong to different groups is easier than that of same group images. We have reported the accuracy obtained for each group in Table~\ref{tab:group}. Also, comparisons with other state-of-the-arts methods are presented in Table~\ref{tab:group}.

\begin{table}[h]
\caption{Comparions of Accuracy (Top-1) with state-of-the-arts}
\label{tab:accuracy}
\begin{center}
\begin{tabular}{|c|c|}
\hline
Algorithm & Accuracy(\%)\\
\hline
\textbf{Proposed Method} & 99.81 \\
\hline
\textbf{Proposed Method with Google Inception} & 99.57\\
\hline
Committe of CNNs \cite{committe} & 99.46\\
\hline
Human Performance \cite{gtsrb} & 98.84\\
\hline
Multi-Scale CNN \cite{multi} & 98.31\\
\hline
pLSA \cite{plsa} & 98.14\\
\hline
Random Forest \cite{rf} & 96.14\\
\hline
\end{tabular}
\end{center}
\end{table}

\section{Conclusion}
 This paper proposes a deep convolutional network with a fewer number of parameters and memory requirements in comparisons to existing methods. The presented network doesn't need data jittering and handcrafted data augmentations. Our main contribution includes the development of modified inception module and a deep network using spatial transformer layer for traffic sign classification.

\section*{Acknowledgment}
I would like to show deep acknowledgement to my professors at IIT Guwahati and IIIT Bangalore for valuable suggestion during this work. Also NVIDIA corporation for donating GPUs for this work.



%

\end{document}